\documentclass[10pt,twocolumn,letterpaper]{article}

\def\arxivpreprint{}
\newif\ifarxiv
\ifdefined\arxivpreprint
  \arxivtrue
\else
  \arxivfalse
\fi

\ifarxiv
  \usepackage[pagenumbers]{wacv}
\else
  \usepackage[review,datasets]{wacv}
\fi

\usepackage{tabularx}
\usepackage{array}
\usepackage{multirow}
\usepackage{siunitx}
\newcommand{\mmreleaseparagraph}{%
  \iftoggle{wacvfinal}{%
    Dataset and code:
    \href{https://huggingface.co/datasets/ckeith/meaningful-moments}{\nolinkurl{huggingface.co/datasets/ckeith/meaningful-moments}},
    \href{https://github.com/SEE-AI-Lab/meaningful-moments}{\nolinkurl{github.com/SEE-AI-Lab/meaningful-moments}}.%
  }{}%
}

\definecolor{wacvblue}{rgb}{0.21,0.49,0.74}
\usepackage[pagebackref,breaklinks,colorlinks,allcolors=wacvblue]{hyperref}

\def\wacvPaperID{74} 
\def\confName{WACV}
\def\confYear{2027}

\title{Meaningful Moments: A Corpus of VLM-Derived Temporal Importance Labels for Video Action Recognition}

\author{Cameron Keith\\
SEE Lab, Dartmouth College\\
{\tt\small cameron.s.keith.26@dartmouth.edu}
\and
SouYoung Jin\\
SEE Lab, Dartmouth College\\
{\tt\small souyoung.jin@dartmouth.edu}
}

\begin{document}
\maketitle
\begin{abstract}
   Not every moment in a video is equally useful for recognizing an action. A few decisive moments often carry much of the evidence: the instant an object moves, the peak of a dive, or the placement of one object on another. Yet existing video datasets rarely provide large-scale, action-conditioned labels identifying which short segments carry recognition evidence. This work asks whether vision--language models can identify meaningful moments at scale and whether the resulting labels improve video action recognition.
   We introduce \emph{Meaningful Moments} (MM), a corpus of about 4.58 million VLM-generated per-segment importance scores for nearly 500{,}000 class-labeled videos spanning Something-Something~v2 (SSv2), Kinetics-400 (K400), and Diving-48 across 622 action classes.
   To evaluate these labels, we freeze the video action classifier and vary only the temporal sampling strategy, using an $\alpha$-parameterized framework that compares importance-led selection against an importance-inverted negative control, separating gains attributable to the VLM importance signal from gains caused by variable-density sampling alone.
   Across all three datasets, VLM-derived importance is informative: at the hard-cut endpoint, keeping VLM-important segments outperforms keeping VLM-designated filler.
   Its downstream usefulness, however, depends strongly on the dataset's class structure: on Diving-48, importance-led fast-forwarding preserves full-video recognition while the inverted control collapses by roughly 30 percentage points; on K400, hard-cut selection beats uniform sampling by 1.45 percentage points, rising to 2.05 under threshold selection; on SSv2, the labels carry signal but do not recover full-video performance.
   Direct scoring matches greedy removal within its noise floor at roughly 25$\times$ fewer oracle calls, and a four-VLM cross-oracle study shows consistent downstream trends.
   We release the importance labels and evaluation protocol under CC-BY-4.0; SSv2 class strings remain under CC-BY-NC-4.0.
\end{abstract}

\section{Introduction}
\label{sec:intro}

\begin{figure*}[t]
\centering
\includegraphics[width=\textwidth]{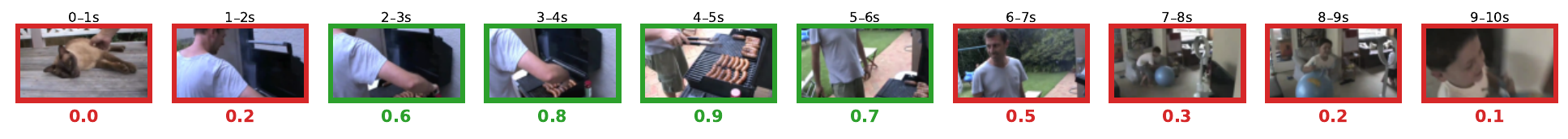}
\caption{A Meaningful Moments example. A K400 test-split clip labeled
\emph{barbecuing} is segmented at $\Delta t = 1.0$\,s; the number under
each segment is the oracle's continuous importance weight, and the
border color marks the binary label (green VLM-important, red
VLM-designated filler). Off-task content (a cat at 0--1\,s, post-action
and indoor scenes at 7--10\,s) receives low weight, while the grilling
action at 2--6\,s forms the kept set.}
\label{fig:intro-teaser}
\end{figure*}

Action recognition depends not only on what appears in a video, but on when the discriminative evidence occurs.
A clip may contain setup, repeated motion, context, or aftermath, while
class-discriminative evidence concentrates in particular temporal
regions: the instant an object moves, the peak of a dive, or the moment
an object is placed on another. Video recognizers are nonetheless
typically evaluated on fixed-length or uniformly sampled clips; uniform
sampling is convenient and often effective, but it cannot adapt when the
relevant evidence is brief, delayed, or concentrated in only part of the
video. Figure~\ref{fig:intro-teaser} shows a concrete example: in a K400
clip labeled \emph{barbecuing}, high importance falls on the grilling
segments and low importance on an unrelated cat and indoor scenes in the
same clip.

Existing datasets and methods address parts of this problem. Action
recognition datasets provide class
labels~\cite{goyal2017something,kay2017kinetics,li2018diving48}; video-language
corpora provide captions or weak language
supervision~\cite{miech2019howto100m,wang2023internvid,chen2024panda70m};
temporal grounding datasets localize query-relevant
moments~\cite{krishna2017activitynet,hendricks2017didemo,lei2021qvhighlights,soldan2022mad};
frame-selection methods learn policies for choosing clips or
frames~\cite{wu2019adaframe,korbar2019scsampler}; and saliency methods
estimate model-specific importance post
hoc~\cite{selvaraju2017gradcam,petsiuk2018rise}. What is missing is a
reusable, large-scale corpus of \emph{action-conditioned} segment
labels: annotations identifying which short temporal segments carry
class-discriminative evidence in class-labeled action videos.

We introduce \emph{Meaningful Moments} (MM), a corpus of approximately
$4.58$~million VLM-generated per-segment importance scores over
nearly $500{,}000$ class-labeled videos and $622$ classes across
SSv2, K400, and Diving-48. Each segment receives a continuous importance
score and a binary pseudo-label, usable as a fine-grained ranking signal
or as a hard kept/dropped decision. Human annotation at this scale would
be expensive and hard to standardize; MM instead uses a
Qwen3-VL-32B oracle~\cite{qwen3vl2025} scoring segments under a
counterfactual minimum-sufficient-subset criterion, conditioned on the
video's action label. These are pseudo-labels, not human ground truth,
and we test their value empirically through downstream recognition. MM
is not generic visual saliency, event-boundary detection, or captioning;
it is action-conditioned temporal evidence labeling, and visually
salient or high-motion segments are not necessarily those that
distinguish the target action.

To make that test controlled, we pair the corpus with an
$\alpha$-parameterized sampling framework
(Sec.~\ref{sec:protocol}). The positive condition samples VLM-important
segments densely and filler sparsely, producing an importance-led
fast-forward; a matched negative control inverts the rule. Both share
their endpoints with uniform full-video sampling, so the experiments
separate gains attributable to the importance signal from gains due to
variable-density sampling itself.

This paper makes three contributions:
\begin{enumerate}
  \item \textbf{Meaningful Moments}, a released large-scale corpus of
    VLM-generated per-segment temporal importance scores and binary
    pseudo-labels for class-labeled videos across SSv2, K400, and
    Diving-48 (Sec.~\ref{sec:corpus});
  \item \textbf{a controlled evaluation protocol}, pairing
    $\alpha$-parameterized variable-density sampling with an
    importance-inverted negative control so that the value of the
    importance signal can be isolated on frozen recognizers
    (Sec.~\ref{sec:protocol});
  \item \textbf{an empirical map of when temporal importance helps}
    (Sec.~\ref{sec:results}): the labels are informative on every
    dataset, but their downstream usefulness is conditional on dataset
    structure, and additional analyses validate the annotation pipeline
    itself (continuous scores dominate the oracle's own kept sets;
    single-call direct scoring matches greedy removal within its noise
    floor at roughly $25\times$ fewer oracle calls; four oracles agree
    moderately yet order K400 and Diving-48 alike).
\end{enumerate}

Together the results support one claim: VLM-derived temporal importance
labels help video action recognition exactly where the oracle's
verification evidence aligns with the evidence the recognizer actually
uses --- an alignment MM makes measurable, at corpus scale, before any
sampler is trained.

\section{Related Work}
\label{sec:related}

\paragraph{Video datasets and temporal annotation.}
Class-labeled action datasets --- SSv2's object interactions, K400's
broad human activities, Diving-48's fine-grained athletic
motion~\cite{goyal2017something,kay2017kinetics,li2018diving48} ---
supply video-level labels: what action is present, not which segments
carry the evidence for it. Video-language corpora scale weak textual
supervision~\cite{miech2019howto100m,wang2023internvid,chen2024panda70m},
but captions describe rather than localize evidence. Temporal grounding
and dense-captioning datasets attach language to temporal
regions~\cite{krishna2017activitynet,hendricks2017didemo,lei2021qvhighlights,soldan2022mad};
QVHighlights is the closest comparison, but its saliency scores are
query-conditioned highlights rather than evidence labels for a known
recognition class. Temporal action localization~\cite{idrees2017thumos,activitynet2015,zhao2017ssn}
marks where action instances begin and end, yet a localized interval
still mixes setup, distractors, and aftermath. MM differs in the object
it annotates: a dense, action-conditioned temporal evidence layer over
the full timeline of class-labeled videos.

\paragraph{Frame, segment, and token selection.}
AdaFrame, SCSampler, and OCSampler learn per-video policies for
selecting frames or clips~\cite{wu2019adaframe,korbar2019scsampler,lin2022ocsampler};
STTS ranks and selects space-time tokens inside video
transformers~\cite{wang2021stts}, EVAD drops tokens for efficient action
detection~\cite{chen2023evad}, and DynamicViT and ToMe prune or merge
tokens in the computation graph~\cite{rao2021dynamicvit,bolya2023tome}.
All treat selection as a learned component inside one system. MM instead
releases the selection \emph{signal} itself --- reusable segment-level
evidence labels that can evaluate or supervise any selector. The frozen
recognizers we test against, VideoMAE and
V-JEPA~2~\cite{tong2022videomae,assran2025vjepa2}, are not
contributions; holding them fixed makes temporal sampling the only
intervention.

\paragraph{VLMs as annotators and judges.}
Foundation-model judgment is established for text (G-Eval, MT-Bench and
LLM-as-judge analyses~\cite{liu2023geval,zheng2023llmasjudge}); MM
extends the model-as-annotator idea to temporally localized evidence
labels for video. Video benchmarks show current VLMs reason over video
while retaining temporal-localization
weaknesses~\cite{fu2024videomme,mangalam2023egoschema}, so we treat
oracle output as pseudo-labels, validated operationally: against an
inverted control, through downstream recognition, and through a
cross-oracle study spanning Qwen3-VL-32B~\cite{qwen3vl2025}, two
proprietary frontier VLMs, and InternVL3-38B~\cite{zhu2025internvl3}.

\paragraph{Importance, saliency, and counterfactual evidence.}
Generic saliency asks what is conspicuous; attribution methods explain
one model's prediction post
hoc~\cite{selvaraju2017gradcam,petsiuk2018rise,ribeiro2016lime,lundberg2017shap,zeiler2014occlusion};
MM targets a third notion, action-conditioned evidence, defined
counterfactually and produced before any downstream model exists.
Removal-based evaluation --- insertion/deletion curves, ROAR,
ROAD~\cite{petsiuk2018rise,hooker2019roar,rong2022road} --- supplies the
validation logic we borrow: a meaningful ordering should degrade
recognition differently when high- versus low-ranked evidence is
removed, with temporal segments rather than pixels as the ranked
objects.

\section{The Meaningful Moments Corpus}
\label{sec:corpus}

\begin{figure}[t]
\centering
\includegraphics[width=\linewidth]{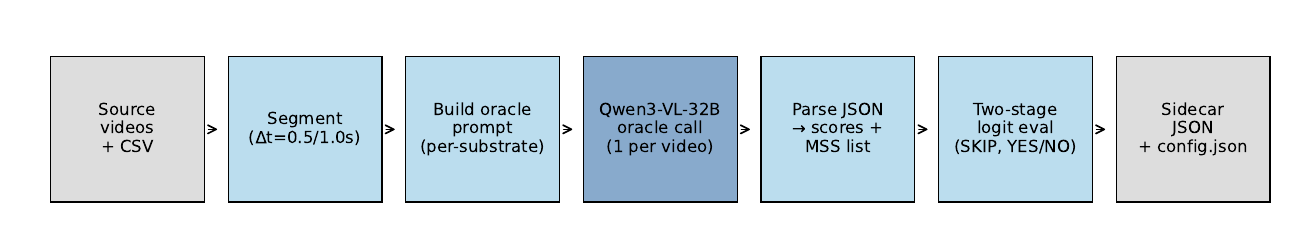}
\caption{The MM construction pipeline. Videos are segmented
on a fixed grid, paired with their class label and a dataset-specific
prompt, and scored in one oracle call. The response yields per-segment scores and a
kept set; a two-stage logit rule yields precheck metadata.}
\label{fig:pipeline}
\end{figure}

Construction pairs each video with its known action label, partitions it
on a fixed temporal grid, and submits it to a Qwen3-VL-32B
oracle~\cite{qwen3vl2025} in one call (Fig.~\ref{fig:pipeline}), which
returns per-segment importance scores, a kept set, and video-level
precheck metadata. The three annotated datasets stress different
temporal structures: SSv2 turns on boundary states, K400's activities
are visible throughout, and Diving-48 concentrates evidence in the dive.

\subsection{Counterfactual definition}
\label{sec:corpus:definition}

Meaningful moments are defined counterfactually: a segment matters to
the extent that the labeled action can no longer be identified without
it. For a video $v$ with label $c$ segmented as
$(s_0, \ldots, s_{T-1})$, the target is a minimum sufficient subset
\[
K^{\star} \in \operatorname*{arg\,min}_{K \preceq v} \lvert K \rvert
\quad \text{s.t.} \quad q(K, c) > \tfrac{1}{2},
\]
where $K \preceq v$ ranges over ordered subsequences of segments and
$q(K, c) = P(\mathrm{YES} \mid \neg\mathrm{SKIP})$ is the oracle's
affirmation probability that $c$ remains identifiable from $K$, computed
from decision-token logits as in Sec.~\ref{sec:corpus:precheck}.

Greedy removal operationalizes this directly, deleting segments one at a
time while the oracle still affirms identifiability. It costs about $25$
oracle calls per eight-segment video and is unstable across runs, so the
production corpus uses \emph{direct scoring}: a single call returning
per-segment scores and an explicit kept set, a one-call approximation of
the same counterfactual target whose faithfulness is an empirical
question answered in Sec.~\ref{sec:results:pipeline}.

\subsection{The operative scoring instruction}
\label{sec:corpus:instruction}

The prompt does not ask an open question about salience. It states the
criterion as a counterfactual test applied segment by segment:
\emph{``If I removed this segment, would the action become harder to
recognize?''} The oracle scores each segment $0$--$100$ under four
banded rubrics: $80$--$100$ means removing the segment would make the
action unrecognizable or ambiguous; $50$--$79$ means it adds meaningful
evidence, showing a distinct phase or disambiguating a near-neighbor
class; $20$--$49$ means mildly helpful but redundant with adjacent
segments; and $0$--$19$ means filler, static pause, scene-only footage,
or no relevant content. The prompt further instructs the oracle to score
by visual content rather than temporal position, to rank segments in
which the action is visibly performed above segments showing only
setting or aftermath, and to err toward inclusion when uncertain. It
then requests an explicit minimum-sufficient-set list of the segments
scoring at least $50$. Dataset-specific variants keep this structure
while tailoring the phase vocabulary; full prompts are in Sec.~D of the
supplement.

\paragraph{Label conditioning and the leakage caveat.}
The action label is an input, not something the oracle infers. Every
call is conditioned on the video's known ground-truth class, so the
oracle \emph{verifies} a label it has already been given rather than
\emph{discovering} what the video depicts. MM scores are therefore
action-conditioned verification evidence, and two consequences hold
throughout this paper. First, the labels must never be read as evidence
the oracle located for an unknown action; a segment's score measures its
contribution to confirming a class supplied in the prompt. Second, a
pipeline without ground-truth labels at annotation time cannot reproduce
this signal directly, and would need an upstream classification step or
an unconditioned prompt, which is a different and untested annotation
target. Conditioning sharpens the signal at the cost of biasing it
toward verification; Sec.~\ref{sec:results:sweeps} shows the downstream
consequence when verification and recognizer evidence diverge.

\subsection{Segments, scores, and binary labels}
\label{sec:corpus:labels}

The timeline is partitioned into $\lceil D/\Delta t \rceil$ exhaustive
segments, with $\Delta t = 0.5$\,s for SSv2 and Diving-48 and $1.0$\,s
for K400 (mean $8.0$, $9.3$, and $10.1$ segments per video, respectively). The integer
score is normalized to a continuous weight $w_i \in [0,1]$, the ranking
signal used by the threshold, budget, and insertion/deletion protocols.
The oracle separately returns an explicit kept-set field from which the
binary label is derived: kept segments are VLM-important, all others
VLM-designated filler. The two fields are not always consistent, so we
treat them as distinct outputs and evaluate both.

\begin{figure}[t]
\centering
\includegraphics[width=\linewidth]{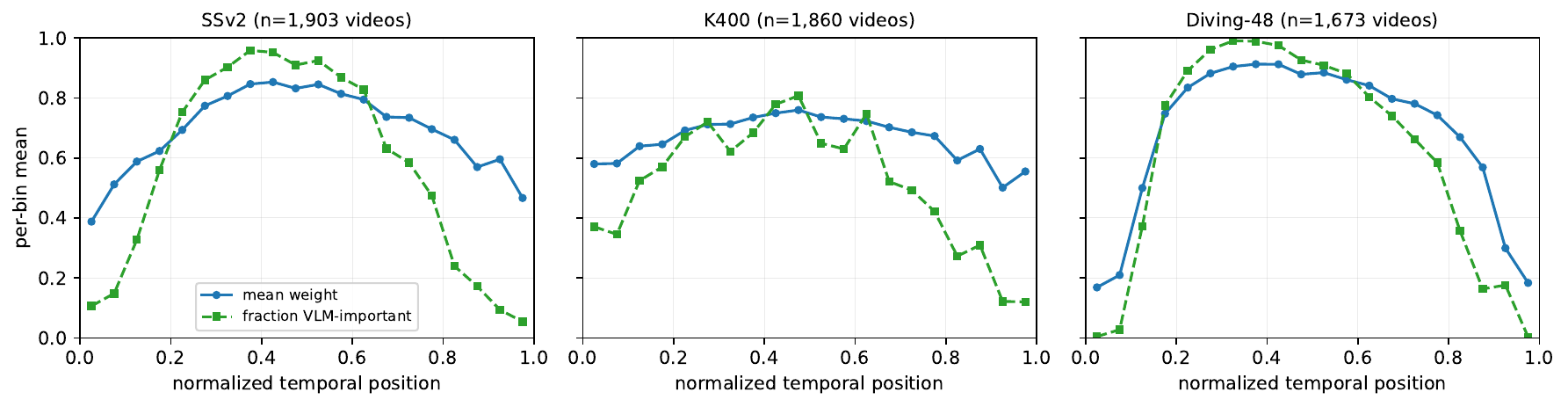}
\caption{Mean continuous weight and fraction labeled VLM-important by
normalized temporal position ($20$ bins, about $2{,}000$ sampled train
videos per dataset). Kept fraction peaks mid-clip and falls toward the
boundaries (first/last bins: SSv2 $0.11$/$0.05$, K400 $0.37$/$0.12$,
Diving-48 $0.00$/$0.00$).}
\label{fig:temporal-profile}
\end{figure}

Figure~\ref{fig:temporal-profile} maps importance onto normalized clip
position. Every dataset concentrates importance mid-clip and depresses
it at the edges, with Diving-48 the most sharply concentrated, K400 the
flattest, and SSv2 falling steeply in the final fifth. This profile is
the direct antecedent of the SSv2 result in
Sec.~\ref{sec:results:sweeps}: importance is depressed precisely in the
boundary fifths where SSv2 classes are discriminated.

\subsection{Precheck and corpus statistics}
\label{sec:corpus:precheck}

Each response also carries a video-level precheck decision (YES, NO, or
SKIP) on whether the labeled action is visible and scorable, computed
from decision-token logits rather than model-written confidence text:
with $\ell_{\mathrm{YES}}, \ell_{\mathrm{NO}}, \ell_{\mathrm{SKIP}}$ the
logits on the three decision tokens,
$P(\mathrm{SKIP}) = \mathrm{softmax}(\ell)_{\mathrm{SKIP}}$ and
$P(\mathrm{YES}\mid\neg\mathrm{SKIP}) =
\sigma(\ell_{\mathrm{YES}}-\ell_{\mathrm{NO}})$. A video passes when
$P(\mathrm{SKIP}) \leq 0.5$ and
$P(\mathrm{YES}\mid\neg\mathrm{SKIP}) > 0.5$. These are visibility
metadata, not calibrated estimates of human correctness.

The released corpus is the precheck-passed subset: $499{,}299$ videos
and $4{,}576{,}082$ per-segment scores across $622$ classes
(Table~\ref{tab:corpus-stats}); with precheck-failed sidecars, which
retain their scores, the total is $4{,}912{,}308$ segments. Parse
failures are zero throughout, so filtering is entirely precheck-driven
and dominated by precheck-NO. Attrition is not uniform: combined
precheck-fail rates are $4.98\%$ on SSv2, $7.80\%$ on K400, and
$15.34\%$ on Diving-48. The corpus and every reported result are
therefore conditional on the oracle's scorability judgment, and
Diving-48 is the most heavily filtered substrate, roughly threefold
over SSv2. MM releases labels, not video content, under CC-BY-4.0
as Parquet tables and per-video JSON sidecars. Source videos retain
their original licenses, and SSv2 class strings retain CC-BY-NC-4.0.
\mmreleaseparagraph

\begin{table}[t]
\centering
\small
\setlength{\tabcolsep}{4pt}
\caption{MM corpus composition. Parse failures are zero throughout, so
``Fail~\%'' is the combined precheck-NO and precheck-SKIP attrition
rate. ``Cls.'' is each dataset's label space; $47$ of Diving-48's $48$
classes appear in the indices used here.}
\label{tab:corpus-stats}
\begin{tabular}{lrrrrrr}
\toprule
Dataset & Attempts & NO & SKIP & Passed & Fail \% & Cls. \\
\midrule
SSv2      & 220{,}847 & 10{,}715 & 290 & 209{,}842 & 4.98  & 174 \\
K400      & 298{,}337 & 23{,}264 & 6   & 275{,}067 & 7.80  & 400 \\
Diving-48 & 16{,}997  & 2{,}577  & 30  & 14{,}390  & 15.34 & 48 \\
\midrule
Total     & 536{,}181 & 36{,}556 & 326 & 499{,}299 & 6.88  & 622 \\
\bottomrule
\end{tabular}
\end{table}

\section{Evaluation Protocol}
\label{sec:protocol}

To measure what the labels are worth, we hold the recognizer fixed and
vary only temporal input sampling. Let $v$ be a video segmented as
$(s_0,\ldots,s_{T-1})$, with oracle weight $w_i \in [0,1]$ and binary
label $\ell_i \in \{\mathrm{important}, \mathrm{filler}\}$ per segment.
Evaluation is video-level: each sampling condition turns the labels into
a sampled recognizer input, and a frozen classifier emits one
prediction. The methodological requirement is to isolate the importance
signal from the sampling mechanism, so the positive and negative
conditions share frame budget, recognizer, video pool, and
preprocessing, and differ only in which segments they treat as dense.

\subsection{Variable-density sampling and the
\texorpdfstring{$\alpha$}{alpha} axis}
\label{sec:protocol:alpha}

The sampler assigns each segment a density $d_i$; effective sampling
mass is duration times density, and frames are allocated across segments
by largest-remainder rounding to hit the recognizer's frame count. For
the positive condition, \emph{Importance-Led FF}, $d_i = 1$ on important
segments and $d_i = \alpha$ on filler. For the matched negative control,
\emph{Anti-Importance FF}, the rule is inverted: $d_i = \alpha$ on
important segments and $d_i = 1$ on filler. Both arms share the same
variable-density machinery; only the importance ordering flips. The
production grid covers interior values
$\alpha \in \{0.05, 0.10, 0.20, \ldots, 0.90\}$, and sampling is
deterministic given a sidecar, condition, and $\alpha$, except where a
protocol is explicitly randomized.

\begin{figure}[t]
\centering
\includegraphics[width=\linewidth]{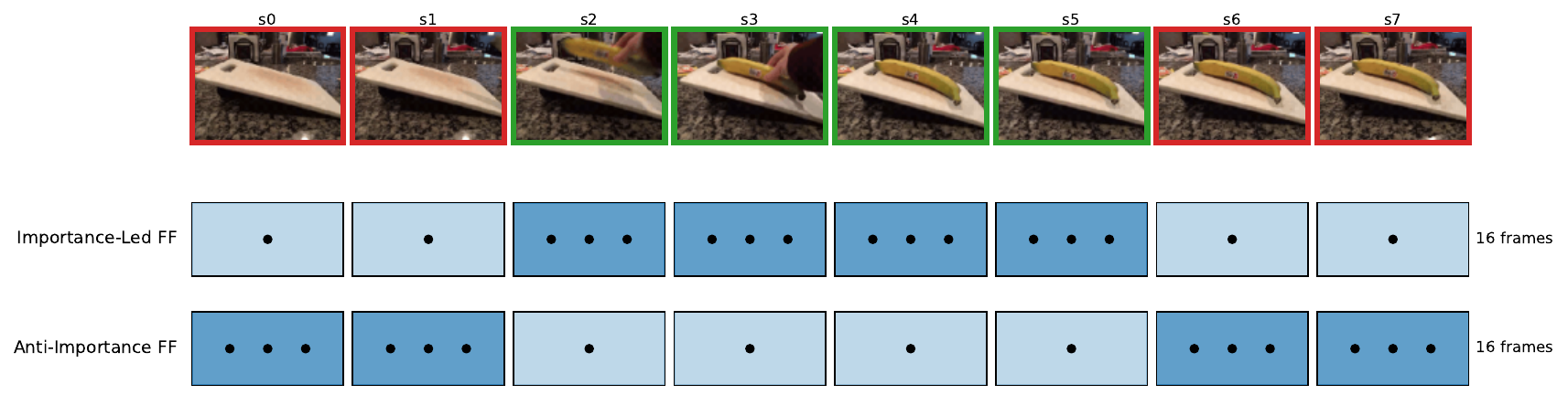}
\caption{Variable-density sampling at $\alpha = 0.25$ on a real
eight-segment SSv2 example with a sixteen-frame budget (production
labels keep $s_2$ through $s_5$). Cell shading shows the per-segment
density $d_i$ and dots show frames allocated by largest-remainder
rounding. Importance-Led FF concentrates frames on VLM-important
segments; the inverted control concentrates the same budget on filler.}
\label{fig:alpha-sampling}
\end{figure}

The matched control is necessary because variable-density sampling can
change recognizer behavior even when the importance signal is
uninformative: it alters which phases are represented and the effective
temporal stride of the input. A positive condition alone cannot separate
VLM-signal gains from sampling-mechanism gains. If gains came purely
from the mechanism, the two curves would behave alike; a gap at matched
$\alpha$ is what indicates task-relevant signal.
Figure~\ref{fig:alpha-sampling} shows both arms on a real eight-segment
SSv2 example.

\paragraph{Endpoint conventions.}
The $\alpha = 1$ reference is \texttt{full}, uniform sampling over the
whole video; it is not a grid point but the shared uniform reference for
both arms. The $\alpha = 0$ endpoint is realized by hard cuts:
\texttt{vlm-selected} (Keep-Important) keeps the important segments and
drops filler, and \texttt{lowest-evidence} (Keep-Filler) inverts that.
Both use cut-and-concatenate followed by resampling rather than a
density rule with $d_i = 0$, so we report them as endpoint-equivalent
conditions expressing the limiting selection rules. A separate
continuous-score endpoint appears in the threshold sweeps:
$T = 0$ keeps every segment with nonzero weight and therefore has
different semantics from the binary hard cut. We keep these conventions
distinct throughout Sec.~\ref{sec:results}.

\subsection{Setup and statistics}
\label{sec:protocol:stats}

Headline experiments use pinned, precheck-passed evaluation pools so
comparisons are paired by video: SSv2 $n = 2{,}080$ over $174$ classes
and K400 $n = 2{,}000$ over $400$ classes, both stratified with seed
$42$, and Diving-48 the full validation split, $n = 1{,}970$ over $47$
classes. Insertion/deletion uses separate $500$-video stratified
subsets. Recognizers are dataset-specific and frozen throughout: no
gradients, no weight updates; only the sampler changes. SSv2 and
Diving-48 use official V-JEPA~2 checkpoints~\cite{assran2025vjepa2}
($16$ frames, $2\times3$ views; $32$ frames, $4\times3$ views), and
K400 uses VideoMAE-L~\cite{tong2022videomae} on a single $16$-frame
clip.

All headline contrasts are paired by video; we report percentile
confidence intervals from $B = 10{,}000$ paired bootstrap resamples,
test paired condition-versus-reference contrasts with McNemar's test
under Yates continuity correction, and apply Bonferroni correction over
the family of $k = 8$ primary contrasts (family-wise
$\alpha = 0.05/8 \approx 0.006$). Videos for which a condition fails to
produce a valid sampled input are dropped from that contrast's matched
subset.

\subsection{Auxiliary protocols}
\label{sec:protocol:aux}

Five further protocols probe robustness and operating-point
sensitivity; each is defined in full in the supplementary material.
\emph{Threshold and budget sweeps} evaluate the continuous weights
instead of the binary kept set, keeping $w_i \ge T/100$ or the
highest-weight segments up to a retained-duration fraction $f$.
\emph{Insertion/deletion curves} rank segments by five orderings
(\texttt{vlm} and \texttt{anti-vlm} weight, \texttt{random},
chronological, and \texttt{motion} flow magnitude) and compare each ordering's trapezoidal
AUC against random~\cite{petsiuk2018rise}. \emph{Temporal-order
shuffling} re-evaluates a kept set with its segments randomly permuted.
\emph{Cross-oracle validation} has four VLMs --- Qwen3-VL-32B, Gemini
3.1 Pro, GPT-5.5, and InternVL3-38B~\cite{zhu2025internvl3} --- score
shared videos on the joint-precheck-pass subset ($n = 369$). A
\emph{closed-set check} has the oracle re-classify SSv2 videos over the
$174$-template closed set, full video against kept-set cut.

\section{Results}
\label{sec:results}

The matched arms separate decisively at the hard-cut endpoint on every
dataset: keeping VLM-important segments beats keeping VLM-designated
filler by $+29.95$\,pp on Diving-48, $+9.04$\,pp on SSv2, and
$+5.25$\,pp on K400. The signal is informative everywhere; its
usefulness depends on dataset structure. Table~\ref{tab:headline}
reports every condition and the best continuous-score operating point:
high-keep selection matches full-video recognition on Diving-48,
aggressive selection significantly improves it on K400, and no tested
operating point recovers full-video performance on SSv2.

\begin{table}[t]
\centering
\footnotesize
\setlength{\tabcolsep}{2pt}
\caption{Headline results, top-1 (\%): the uniform reference, both
hard-cut endpoints, and the best threshold and budget operating points
with the winning $T$ or $f$ in parentheses. $\Delta$ is best minus
\texttt{full}; of the three, only the K400 entry survives Bonferroni
correction (Table~\ref{tab:cis}). \texttt{v-sel} and \texttt{low-ev} abbreviate
\texttt{vlm-selected} and \texttt{lowest-evidence}.}
\label{tab:headline}
\begin{tabular}{l S S S c c S}
\toprule
{Dataset} & {\texttt{full}} & {\texttt{v-sel}} & {\texttt{low-ev}} & {Best $T$} & {Best $f$} & {$\Delta$} \\
\midrule
Diving-48 & 87.82 & 82.08 & 52.13 & 88.58 (0)  & 87.87 (90) & {$+0.76$} \\
SSv2      & 65.77 & 62.45 & 53.41 & 63.37 (50) & 63.65 (70) & {$-2.12$} \\
K400      & 83.95 & 85.40 & 80.15 & 86.00 (70) & 85.75 (30) & {$+2.05$} \\
\bottomrule
\end{tabular}
\end{table}

\begin{table}[t]
\centering
\footnotesize
\setlength{\tabcolsep}{3pt}
\caption{Paired-bootstrap CIs on all eight primary contrasts, each
against \texttt{full}, from $B = 10{,}000$ resamples. ``$\ast$'' marks
intervals excluding $0$ at the raw $95\%$ level, ``$\dagger$'' those
additionally surviving Bonferroni correction over the $k = 8$ family.
D48 $T{=}0$ is the one raw-significant contrast that does not survive;
the two unmarked D48 rows were never significant.}
\label{tab:cis}
\begin{tabular}{l r r l}
\toprule
{Contrast} & {$n$} & {$\Delta$ (pp)} & {$95\%$ CI} \\
\midrule
\multicolumn{4}{l}{\emph{$\alpha$-sweep}} \\
D48 Imp-Led $\alpha{=}0.70$ & 1970 & $+0.56$               & $[-0.10, +1.27]$ \\
SSv2 \texttt{vlm-selected}  & 2080 & $-3.32^{\ast\dagger}$ & $[-4.90, -1.73]$ \\
K400 \texttt{vlm-selected}  & 2000 & $+1.45^{\ast\dagger}$ & $[+0.35, +2.55]$ \\
\midrule
\multicolumn{4}{l}{\emph{$\alpha = 0$ continuous-score endpoints}} \\
D48 $T{=}0$     & 1970 & $+0.76^{\ast}$        & $[+0.05, +1.47]$ \\
SSv2 $T{=}50$   & 2080 & $-2.40^{\ast\dagger}$ & $[-3.75, -1.01]$ \\
K400 $T{=}70$   & 2000 & $+2.05^{\ast\dagger}$ & $[+1.00, +3.10]$ \\
D48 $f{=}90\%$  & 1970 & $+0.05$               & $[-0.66, +0.76]$ \\
K400 $f{=}30\%$ & 2000 & $+1.80^{\ast\dagger}$ & $[+0.65, +2.90]$ \\
\bottomrule
\end{tabular}
\end{table}

Every SSv2 and K400 contrast is significant at the raw $95\%$ level and
survives Bonferroni correction, so the SSv2 kept-set deficit
($-3.32$ and $-2.40$\,pp) and the K400 gains ($+1.45$, $+2.05$,
$+1.80$\,pp) are robust. The Diving-48 contrasts are not: the
$T = 0$ win does not survive correction and the $\alpha = 0.70$ and
$f = 90\%$ contrasts were never significant, so the Diving-48 headline
reads as selection \emph{matching} \texttt{full}, not beating it. On
SSv2 the CI-tested threshold contrast is $T = 50$ at $-2.40$\,pp; the
$-2.12$\,pp figure in Table~\ref{tab:headline} is the best raw operating
point on the budget axis and carries no significance claim.

\subsection{Alpha sweeps}
\label{sec:results:sweeps}

\begin{figure*}[t]
\centering
\includegraphics[width=0.32\textwidth]{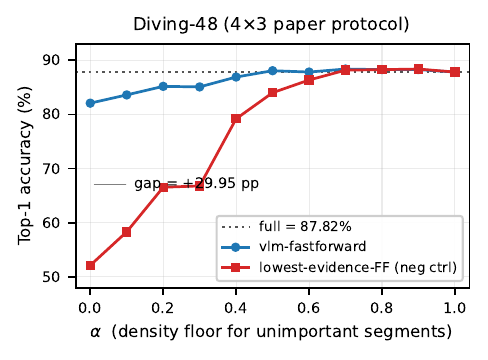}\hfill
\includegraphics[width=0.32\textwidth]{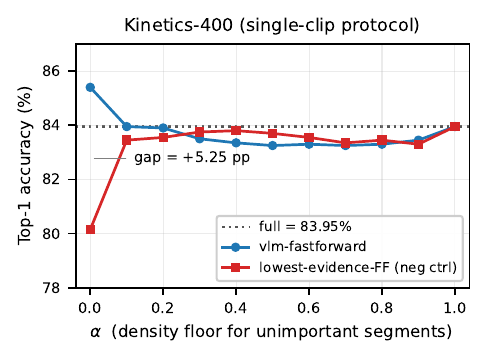}\hfill
\includegraphics[width=0.32\textwidth]{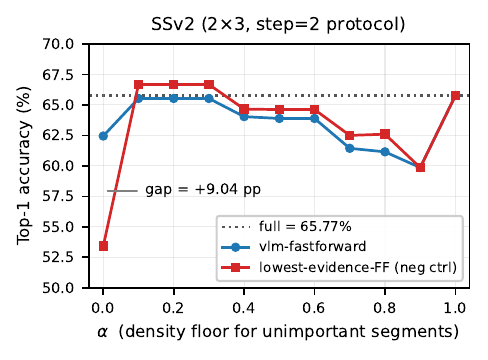}
\caption{Top-1 versus $\alpha$ for Diving-48 (left), K400 (center), and
SSv2 (right), Importance-Led FF against the matched inverted control.
Diving-48 shows the intended signature, the arms fanning out as
$\alpha \to 0$; K400's interior is flat; SSv2's interior curves overlap,
the control marginally above, a protocol artifact discussed in the
text. Decile tables are in the supplementary material.}
\label{fig:alpha-curves}
\end{figure*}

\paragraph{Diving-48: fast-forward works when evidence is concentrated.}
Here the mechanism behaves as designed (Fig.~\ref{fig:alpha-curves}).
The arms agree near $\alpha = 1$ and fan out as $\alpha \to 0$, the
signature of the importance signal rather than the sampling mechanism
driving the gap. Importance-Led FF stays within $\pm 0.6$\,pp of
\texttt{full} across $\alpha \in [0.5, 1.0]$ and peaks at
$\alpha = 0.7$ ($+0.56$\,pp, CI overlapping zero, so suggestive rather
than significant), while the inverted control collapses monotonically to
$52.13\%$. A $+29.95$\,pp endpoint gap is not something
variable-density sampling alone can produce.

\paragraph{K400: hard cuts beat uniform sampling.}
The interior of the $\alpha$ axis is uninformative --- every interior
gap satisfies $\lvert\text{gap}\rvert \le 0.5$\,pp --- because K400
activities are visible through most of the clip, so a uniform sample
already covers the action and there is little filler to skip. The hard
cut instead discards setup and camera-pan segments outright and spends
the sixteen-frame budget on footage the oracle scored as carrying the
strongest verification evidence. Keep-Important is the best K400
hard-cut condition at $85.40\%$, significantly beating \texttt{full} by
$+1.45$\,pp ($[+0.35, +2.55]$, survives Bonferroni) and Keep-Filler by
$+5.25$\,pp.

\paragraph{SSv2: informative, but full context still wins.}
Importance-Led FF never exceeds \texttt{full} at any $\alpha$; its best
interior point is $0.24$\,pp below. The interior gaps are small and run
in the \emph{wrong} direction: the arms tie at $\alpha = 0.9$, and below
it the inverted control is marginally ahead (gaps of $-0.62$ to
$-1.45$\,pp), even edging the \texttt{full} baseline at
$\alpha \le 0.3$ ($66.68$ versus $65.77$). That last comparison is a
protocol artifact, not a control win. We report the primary
\texttt{frame\_step=2} protocol because the step=4 paper protocol
saturates the variable-density pool on short SSv2 videos; the step=2
reference sits about $4$\,pp below its step=4 counterpart ($65.77$
versus $69.76$), and interior variable-density inputs alter the
effective temporal stride. The hard cut removes the confound and gives a
clean $+9.04$\,pp gap ($+11.77$\,pp under step=4). The reversal traces
to misalignment: SSv2 classes are discriminated by initial versus final
object state, so the oracle tags mid-motion as important and the
near-still boundary frames the recognizer wants as filler.

\subsection{Continuous scores dominate the kept set}
\label{sec:results:continuous}

The oracle returns both a continuous weight and a categorical kept set,
and they are not equally useful: the kept set is dominated by other
operating points on all three datasets (Table~\ref{tab:headline}; full
sweeps in the supplementary material). Three shapes emerge. Diving-48 is
endpoint-optimal at $T = 0$, which keeps every segment with nonzero
weight; because $T = 0$ and \texttt{full} are both keep-everything
conditions under different inference protocols, that $+0.76$\,pp is best
read as the kept-set protocol matching the uniform protocol rather than
as a selection gain. SSv2 is interior-optimal on both axes with the
whole sweep below \texttt{full}, and K400 peaks above \texttt{full} in
the aggressive regime, adding $+0.60$\,pp over the kept set. Two
observations hold everywhere. The kept set is never optimal: the oracle
is stricter than its own scoring rule prescribes, omitting roughly
$12\%$ of segments scoring at or above its stated $50/100$ bar
(measured on Diving-48). And the low-weight tail is not redundant ---
forcing $T = 100$ or $f = 10$ costs accuracy on every dataset, falling
below even the Keep-Filler control on SSv2 and Diving-48.

\subsection{Which SSv2 templates lose}
\label{sec:results:templates}

A recognizer-independent check localizes the SSv2 deficit. The oracle
itself re-classified the SSv2 test set ($n = 25{,}647$) over the full
$174$-template closed set, full video against kept-set cut; top-1 falls
from $17.18\%$ to $14.70\%$, so the cut preserves about $85\%$ of
full-video top-1 from the oracle's own perspective. The loss is not
uniform: over the $172$ templates with at least $10$ videos the median
delta is exactly $0$\,pp and $67\%$ sit within $\pm 2$\,pp of full,
while $15\%$ lose more than $5$\,pp and $2\%$ gain more than $5$\,pp.
The split runs along exactly the boundary-versus-motion line the
$\alpha$-sweep predicts (Table~\ref{tab:templates}).

\begin{table}[t]
\centering
\footnotesize
\setlength{\tabcolsep}{3pt}
\caption{SSv2 per-template split under kept-set closed-set
classification (templates with $n \ge 10$). Positive $\Delta$ means full
beats the cut. Losses are boundary-state-discriminated, their evidence
being the clip-boundary state that cut-and-concatenate repacking
disturbs; gains are motion-discriminated.}
\label{tab:templates}
\begin{tabular}{@{}p{0.44\linewidth}rrrr@{}}
\toprule
{Template} & {$n$} & {Full} & {Cut} & {$\Delta$ (pp)} \\
\midrule
\multicolumn{5}{@{}l}{\emph{Largest losses: boundary-state}} \\
Removing something, revealing something behind &  77 & 36.36 & 15.58 & $+20.78$ \\
Dropping something next to something           & 106 & 36.79 & 16.04 & $+20.75$ \\
Putting something on a surface                 & 492 & 51.83 & 31.10 & $+20.73$ \\
Laying something on the table on its side      & 145 & 35.86 & 16.55 & $+19.31$ \\
\midrule
\multicolumn{5}{@{}l}{\emph{Largest gains: motion}} \\
Something falling like a feather or paper      & 358 & 13.97 & 22.35 & $-8.38$ \\
Tearing something into two pieces              & 581 & 36.32 & 41.65 & $-5.34$ \\
Pouring something out of something             & 119 & 31.09 & 35.29 & $-4.20$ \\
Throwing something against something           &  88 &  1.14 &  4.55 & $-3.41$ \\
\bottomrule
\end{tabular}
\end{table}

\subsection{Mechanism probes}
\label{sec:results:mechanism}

Insertion/deletion curves test the importance \emph{ordering} across the
whole retained-fraction axis rather than at one operating point. Two
results hold across datasets: inverting the signal hurts everywhere
(\texttt{anti-vlm} is below \texttt{random} by $0.098$ AUC on SSv2,
$0.032$ on K400, and $0.070$ on Diving-48, all $p < 10^{-3}$), and
\texttt{vlm} beats \texttt{random} significantly on top-5 on all three
($+0.015$, $+0.013$, $+0.068$) while on top-1 it is significant only on
Diving-48 ($+0.054$): the ordering reaches the right class neighborhood
more reliably than it nails the exact label.

\paragraph{Is the oracle worth its cost against optical flow?}
Optical-flow magnitude is the cheapest content-aware alternative to a
VLM, and the comparison is mixed rather than uniformly favorable. On
SSv2 insertion/deletion, \texttt{motion} edges the oracle on top-1
($+0.0164$ versus $-0.0002$ $\Delta$AUC against random; the flow
interval excludes zero; the VLM interval does not), the ordering-level
counterpart of the boundary-state misalignment above. On K400 and
Diving-48 the ranking reverses: the oracle leads flow by $+0.016$ top-1
$\Delta$AUC on K400 ($+0.0078$ versus $-0.0082$) and by $+0.008$ on
Diving-48 ($+0.0538$ versus $+0.0460$), and it leads flow on top-5 AUC
on all three, which is where its finer class information shows up. The
cost asymmetry is stark: flow is free and needs no labels, while MM cost
about $0.54$M oracle calls. The defensible reading is that the VLM
signal is not reducible to motion energy and carries class-level
information flow does not, but it does not dominate flow uniformly, and
where a recognizer's evidence is boundary-state rather than mid-action a
flow baseline is competitive at a fraction of the cost. A practitioner
with a fixed budget should check which regime their dataset is in before
paying for oracle calls --- which is exactly what the corpus-level
temporal profile of Sec.~\ref{sec:corpus:labels} makes checkable.

Temporal-order shuffling shows dataset-specific rather than
selection-specific reliance. SSv2 conditions are both strongly
order-reliant, K400 conditions negligibly so, and on Diving-48 the sign
reverses significantly: \texttt{vlm-selected} is \emph{more}
order-reliant than \texttt{uniform} (differential $-5.59$\,pp,
$[-8.09, -3.15]$), because the kept set concentrates on the dive whose
counted rotations shuffling destroys.

\subsection{Pipeline validation}
\label{sec:results:pipeline}

\paragraph{Cross-oracle agreement.}
On the $n = 369$ joint-precheck-pass sample, agreement has a clear
three-versus-one structure. The three frontier-tier models
(Qwen3-VL-32B, Gemini 3.1 Pro, GPT-5.5) cluster at pairwise Spearman
$0.49$--$0.61$ on per-segment importance, while InternVL3-38B sits apart
at $0.35$--$0.37$ against all three. That the outlier is consistent
against every member of the trio, rather than specific to one pair,
suggests the importance prior tracks model family and capability tier
rather than a single comparison's artifact. Kept-set agreement is
moderate everywhere (Set-F1 $0.55$--$0.74$). The load-bearing result is
downstream: substituting any other oracle preserves the qualitative
ordering, with \texttt{vlm-selected} beating \texttt{uniform} and
\texttt{lowest-evidence} on Diving-48 and K400 for every oracle.

\paragraph{Direct scoring versus greedy removal.}
Three pilot results on the same $101$ SSv2 videos justify approximating
greedy removal with a single call. Greedy is itself unstable: four runs
with identical prompt, model, and parameters agree pairwise at only mean
Jaccard $0.146$, and its apparently stable rank signal is a tie-handling
artifact, dropping to about $0.17$ Spearman under tie correction. Pooled
greedy-to-direct agreement is $0.144$, the same magnitude as greedy's
own noise floor, so switching modes moves the kept set no more than
re-running greedy would while cutting calls from roughly $25$ per video
to one --- $0.54$M against an estimated $13.4$M, which at fixed budget
would have supported only about $21$K labeled videos. Extraction mode
rather than prompt wording is load-bearing: two direct runs differing
only in prompt agree at Jaccard $0.827$ and Spearman $0.861$, while
either agrees with greedy at only $0.21$ Jaccard.

The full closed-set protocol, per-class diagnostics, extended sweep
tables, and a frozen-encoder adaptation diagnostic separating domain
shift from irrecoverable information loss are in the supplementary
material.

\section{Discussion}
\label{sec:discussion}

\paragraph{What an MM score means.}
An MM importance score measures how much a segment contributes to a
verification-oriented VLM's ability to confirm the labeled action:
counterfactual in origin, action-conditioned by construction,
approximated by a single direct-scoring call. The results show this
captures something real --- inverting it hurts recognition on every
dataset, and the signal is not reducible to motion --- but it does not
promise identity with \emph{recognizer} importance. SSv2 exposes the gap
most clearly: the oracle's verification evidence concentrates in
mid-action motion while a frozen recognizer discriminates SSv2 classes
by boundary states, and that misalignment (together with the step-2
baseline artifact noted in Sec.~\ref{sec:results}) produces the reversed
interior-$\alpha$ gaps. The score is what a VLM needs to verify the
labeled action, which only partly matches what a recognizer needs
to classify the video.

\paragraph{Why dataset structure decides whether selection helps.}
Two conditions matter: \emph{alignment} between what the oracle marks
and what the recognizer uses, and \emph{concentration} of that evidence
in a small part of the clip. Diving-48 satisfies both --- its classes
are stereotyped temporal arcs whose discriminative content is the dive
itself --- so selection preserves recognition, and the shuffle reversal
confirms that the kept set concentrates order-defining rotation content.
K400 satisfies alignment without concentration: its appearance-cued
classes are visible throughout, so the interior of the $\alpha$ axis is
flat, and the hard-cut gain comes from spending a fixed sixteen-frame
budget on unambiguous footage rather than from finding hidden evidence.
SSv2 satisfies concentration without alignment, and selection stays
below \texttt{full}. Dataset difficulty is not the relevant variable; where
the discriminative evidence lives relative to what a
verification-oriented oracle marks is --- and because the corpus-level
temporal profile makes that relationship visible, it can be reasoned
about \emph{before} running a recognizer. The frozen-encoder
head-adaptation diagnostic (supplementary) further suggests that much of
the off-the-shelf kept-set deficit is closeable domain shift rather than
missing evidence.

\paragraph{What oracle disagreement means.}
Frontier-tier oracles agree at Spearman $0.49$--$0.61$ while
InternVL3-38B sits apart at $0.35$--$0.37$ against all three. A
single-oracle MM label is one competent annotator's judgment, not ground
truth. The reliability unit is the aggregate: per segment, oracles
disagree; pooled over Diving-48 and K400, every oracle reproduces the
same recognition ordering. Applications needing per-segment identity
should ensemble oracles or add human verification.

\paragraph{Operational guidance for VLM-as-oracle pipelines.}
Measure the expensive extraction mode's own test-retest noise first ---
greedy removal agreed with itself at only $0.146$ Jaccard, so a
$25\times$ cheaper single-call mode matching it within that floor is
strictly preferable. Extraction mode dominates prompt choice (cross-mode
kept-set agreement $0.21$ versus $0.83$ across prompts within direct
mode). Ask for continuous scores and derive categorical decisions
downstream.

\paragraph{Future work.}
The binary labels can supervise temporal-selector
training~\cite{wu2019adaframe,wang2021stts}, the continuous weights
can reward learned sampling policies, and a human-calibrated
importance benchmark is the natural next annotation.

\section{Limitations}
\label{sec:limitations}

MM has no human temporal-importance ground truth; cross-oracle agreement
bounds reliability only from above. Scoring conditions on the known
action label, so the labels are verification evidence and cannot support
discovery claims. Results come from one frozen recognizer per dataset
(V-JEPA~2 on SSv2/Diving-48, VideoMAE-L on K400), and the kept-set
boundary inherits the oracle's prompt-defined thresholding behavior.
Precheck filtering passes different fractions per dataset (attrition
$15.34\%$ on Diving-48, $7.80\%$ on K400, $4.98\%$ on SSv2), so corpus
coverage is biased toward oracle-scorable videos --- most strongly on
the dataset with the largest hard-cut gap. Importance labels are
released under CC-BY-4.0, SSv2 class strings remain under
CC-BY-NC-4.0, and source videos retain their original licenses.

\section{Conclusion}
\label{sec:conclusion}

Meaningful Moments releases $4.58$M action-conditioned temporal
importance labels over $499$K videos. The labels are informative on
every dataset we annotate, but they improve recognition only where
the oracle's verification evidence aligns with the evidence the
recognizer actually uses. MM makes that alignment measurable at
corpus scale, before any sampler is trained.

{
    \small
    \bibliographystyle{ieeenat_fullname}
    \bibliography{main}
}

\end{document}